%% file: main.tex
\documentclass[10pt,twocolumn,letterpaper]{article}

\usepackage{iccv}
\usepackage{times}
\usepackage{epsfig}
\usepackage{graphicx}
\usepackage{amsmath}
\usepackage{amssymb}
\usepackage{booktabs} % For formal tables
\usepackage[breaklinks=true,bookmarks=false]{hyperref}

\iccvfinalcopy % *** Uncomment this line for the final submission

\def\iccvPaperID{12686} % *** Enter the ICCV Paper ID here

\ificcvfinal\fi

\begin{document}

%%%%%%%%% TITLE
\title{ Make-An-Animation: \\ Large-Scale Text-conditional 3D Human Motion Generation}

\author{Samaneh Azadi
\and
Akbar Shah
\and Thomas Hayes\\
Meta AI
\and Devi Parikh
\and Sonal Gupta 
\\
}

\maketitle
\ificcvfinal\thispagestyle{empty}\fi

%%%%%%%%% ABSTRACT
\begin{abstract}
Text-guided human motion generation has drawn significant interest because of its impactful applications spanning animation and robotics. Recently, application of diffusion models for motion generation has enabled improvements in the quality of generated motions. However, existing approaches are limited by their reliance on relatively small-scale motion capture data, leading to poor performance on more diverse, in-the-wild prompts. In this paper, we introduce Make-An-Animation, a text-conditioned human motion generation model which learns more diverse poses and prompts from large-scale image-text datasets, enabling significant improvement in performance over prior works. Make-An-Animation is trained in two stages. First, we train on a curated large-scale dataset of (text, static pseudo-pose) pairs extracted from image-text datasets. Second, we fine-tune on motion capture data, adding additional layers to model the temporal dimension. Unlike prior diffusion models for motion generation, Make-An-Animation uses a U-Net architecture similar to recent text-to-video generation models. Human evaluation of motion realism and alignment with input text shows that our model reaches state-of-the-art performance on text-to-motion generation. Generated samples can be viewed at \href{https://azadis.github.io/make-an-animation}{https://azadis.github.io/make-an-animation}.

\end{abstract}

\begin{figure}
  \centering \includegraphics[width=0.49\textwidth]{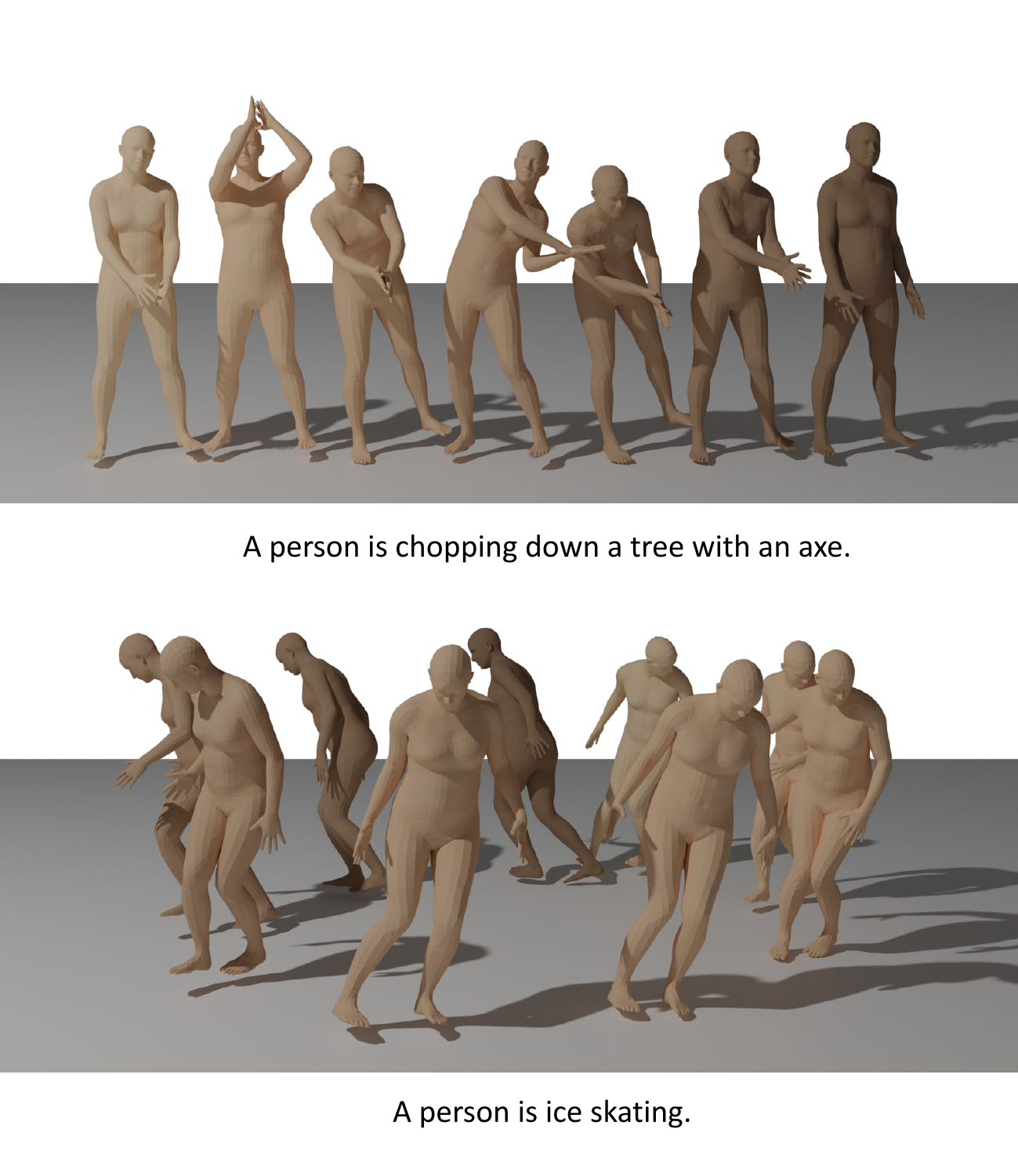}
  \caption{Samples generated by Make-An-Animation for text conditional motion generation. The lighting of the body models represents progress across time. Darker color indicates later frames in the sequence. In the top image, for a better visualization, frames are distributed horizontally. }
  \label{fig:teaser}
\end{figure}
%%%%%%%%% BODY TEXT
\section{Introduction}
As the world shifts towards virtual spaces, and as embodied agents become more capable, it will be increasingly important to be able to generate plausible human motion. Speech or text are perhaps the most natural ways to prompt generative models, which is one reason behind the explosion in text-to-x generation research. Human motion generation from text has diverse applications both in virtual and real worlds. For instance, it enables control of robots with speech, faster video game development, creation of special effects featuring humans, as well as novel metaverse user-interaction modes whereby users can control the actions of their or others' avatars via voice or text. 

Text-conditioned human motion generation is challenging for a number of reasons. Firstly, there exists a huge diversity of human motions and text descriptions which describe those motions (i.e., the task is many-to-many distribution matching). Secondly, because motion capture data is expensive to acquire, human motion datasets are limited in scale and diversity. 

Recently, text-to-motion generation has seen rapid improvements in the quality and diversity of poses generated. In particular, diffusion models, which have yielded impressive results in text-conditioned image and video generation, have proven effective in generating human body poses \cite{tevet2022mdm, kim2022flame, zhang2022md}. However, these models struggle with in-the-wild prompts that are outside of the distribution of motion capture data. 

To address these shortcomings, we present Make-An-Animation, a text-conditioned human motion generation model which learns to map diverse human poses to natural language descriptions through in-the-wild image-text datasets and significantly improves on prior state-of-the-art models which only rely on motion capture data.

Core to our approach is a large-scale dataset of human poses extracted from image-text datasets. Our key observation is that we should not be limited to learning human motion from motion capture data when we have access to large-scale in-the-wild video and images of human poses. So, to address the limited scale and diversity of motion capture datasets, we extract a large-scale Text Pseudo-Pose (TPP) dataset from image-text datasets filtered for images containing humans. This TPP dataset contains 35M (text, static pose) pairs. %Additionally, this approach of extracting pseudo-pose labels from in-the-wild image or video datasets opens the door for large-scale modeling of human interactions with objects and scenes.

Make-An-Animation is trained in two stages: (1) We train a text-conditioned static 3D pose generation diffusion model on the TPP dataset. In this stage, Make-An-Animation learns the distribution of human poses and their alignment with text. (2) Then, we extend the pre-trained diffusion model to motion generation via the addition of temporal convolution and attention layers which model the new temporal dimension and train on widely-used motion capture data. In this second stage, the model learns motion, i.e., how to connect poses in a temporally coherent manner. Crucially, it does not have to re-learn the distribution of feasible poses or their alignment with text. 

The Make-An-Animation architecture is a U-Net, similar to recent text-to-video diffusion models. We condition the U-Net on text representations extracted from a language model trained on large-scale language data. We represent human motion as a sequence of 3D SMPL body parameters, with a 6D continuous SMPL representation for 21 body joints and the root orient. We additionally represent the global position per frame via a 3D vector indicating the position in each of the $x, y, z$ dimensions.

Through human evaluation on a collected set of 400 diverse text prompts, we demonstrate that our method outperforms prior works in terms of generated pose realism and text alignment. 

To summarize, our main contributions are:
\begin{itemize}
    \item We present Make-An-Animation -- a text-conditioned human motion generation model which improves on prior state-of-the-art models, especially on diverse, in-the-wild text prompts.
    \item We show, for the first time, how to leverage large-scale image datasets to learn in-the-wild human poses for generation. We show through ablations that pre-training on our collected Text Pseudo-Pose dataset significantly improves the generalization to prompts outside the distribution of widely-used motion capture datasets while showing a comparable performance on the mocap test set.
    \item We present a U-Net architecture for human motion generation which leverages a language model pre-trained on large-scale language data and naturally extends a static text-to-pose generation diffusion model to motion generation via the addition of temporal convolution and attention layers. 
\end{itemize}

\begin{figure*}
  \centering \includegraphics[width=\textwidth]{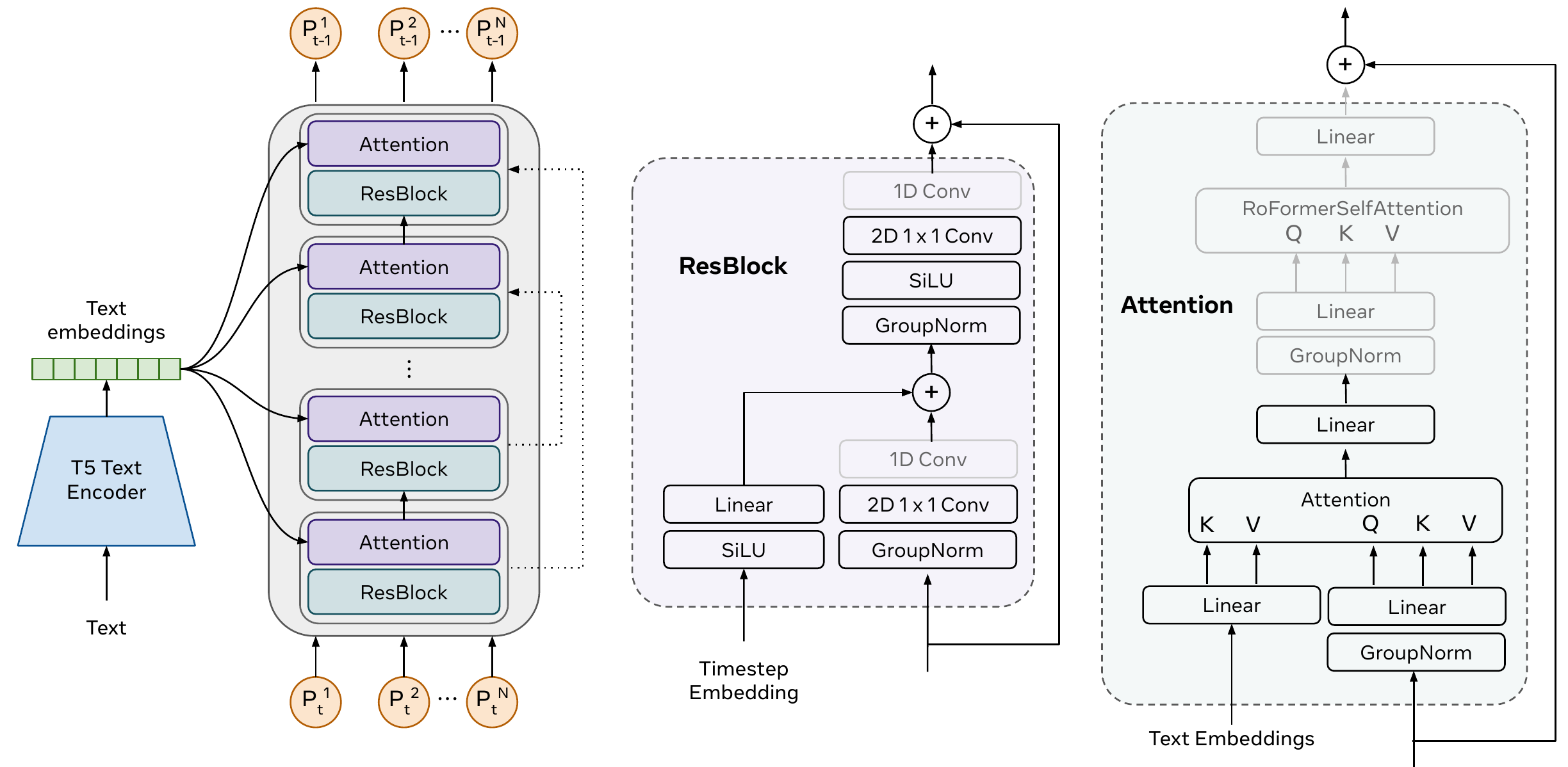}
  \caption{\textbf{Make-An-Animation Model Architecture.} Our diffusion model is built on a U-Net architecture inspired by recent image and video generation models. The U-Net consists of a sequence of Residual Blocks with 1x1 2D-convolution layers and Attention Blocks with cross-attention on textual information. To model the temporal dimension, we add 1D temporal convolution layers after each 1x1 2D-convolution, as well as temporal attention layers after each cross-attention layer. These temporal layers (greyed out in the figure) are only added in the motion fine-tuning stage.}
  \label{fig:model_arch}
\end{figure*}

\section{Related Works}

 \paragraph{Human pose and motion generation}
 
Prior works in human pose and motion synthesis have explored generative models either unconditionally~\cite{yan2019, zhao2020, modi2022}, or conditioned on various input signals such as a prior motion~\cite{julieta2017, ruiz2019}, an action class~\cite{guo2020a2m, petrovich2021actor, cervantes2022}, or music~\cite{lee2019, li2021}. 

Using text descriptions as a guidance in pose and motion synthesis has been a more recent research direction where many existing works use 2D keypoints as pose representations. \cite{zhou2019pose} selects a base pose from 8 clusters based on an input text fed to a GAN model to generate a human image. \cite{zhang2021} uses a GAN to generate a set of heatmaps for body keypoints conditioned on the text input. \cite{roy2022} proposes a coarse-to-fine approach where a refinement stage is introduced on top of the initial coarse estimate of the keypoint heatmaps. Different from the above methods, \cite{briq2021} uses SMPL to represent a 3D body pose, generated by an LSTM GAN from a text input.

 Many existing works approach text-to-motion by learning to align text and pose or motion embeddings in the feature space. JL2P~\cite{ahuja2019jl2p} proposes to learn the joint embedding of text and pose using an autoencoder with curriculum learning. \cite{ghosh2021} proposes a two-stream model to encode upper and lower body motions separately. AvatarCLIP~\cite{hong2022avatarclip} uses a pre-trained VPoser model to generate candidate poses, which are then used to optimize a motion VAE. MotionCLIP~\cite{tevet2022motionclip} trains an auto-encoder while simultaneously reconstructing motion and aligning the motion manifold with CLIP's latent space. TEMOS~\cite{petrovich2022temos} trains a joint latent space through separate text and motion transformer encoders, allowing non-deterministic motion sampling. T2M~\cite{guo2022t2m} also uses a VAE model, but encodes motion as snippets and introduces an extra sampling for motion length conditioned on the input text.

Leveraging recent advancements in diffusion models, Motion Diffusion Model (MDM)~\cite{tevet2022mdm}, MotionDiffuse~\cite{zhang2022md}, and FLAME~\cite{kim2022flame} adopt a diffusion process from text to motion with a transformer, which can significantly improve the diversity of synthesized samples. MDM and MotionDiffuse are trained on HumanML3D~\cite{guo2022t2m} which partially limits their choice of representation to stick figures for text-to-motion generation, while our curated large-scale dataset provides millions of SMPL pose labels. MDM supports converting poses from stick figures to SMPL bodies only through an optimization SIMPLIFY-X\cite{simplify-x} procedure that often results in unrealistic body poses.

\paragraph{Diffusion generative models} 
Diffusion models are a class a generative model based on the thermodynamic stochastic diffusion process. In the forward diffusion process, a sample is drawn from the data distribution and is gradually noised by a diffusion process. A neural network learns the reverse process to gradually denoise the sample. 

Diffusion models have recently enabled rapid advancements in image generation. \cite{ddpms} first demonstrated diffusion models for unconditional image generation. In comparison to GANs ~\cite{goodfellow2014} and VAEs, diffusion models have shown superior training stability, avoid mode dropout, and yield stronger performance. Adapting diffusion models for conditional generation, \cite{diff_beat_gans} introduced classifier-guided diffusion. More recently, classifier free guidance~\cite{cfg} has been introduced to enable conditioning without the need for a separate classifier. Imagen~\cite{Imagen} showed that conditioning on text embeddings extracted from a pre-trained language model enhanced text alignment, and that performance improved with text encoder size. Our U-Net architecture shares similarities with Imagen. Namely, Make-An-Animation uses the same text encoder, T5-XXL~\cite{t5_2020}, and a similar U-Net architecture. Other improvements have been suggested to improve training stability and performance, such as learning the reverse diffusion variances \cite{nichol2021improved} and v-parameterization~\cite{progdistill}. Our method leverages both of these improvements.

\section{Text-to-3D Human Motion Generation}

\subsection{Dataset}
\label{sec:mocap}

\paragraph{3D Human motion datasets.} We use the AMASS dataset of 3D human motions~\cite{AMASS:ICCV:2019} and its textual annotations from the HumanML3D dataset~\cite{HumanML3d}. We borrow the original SMPL annotations from AMASS instead of the processed data from HumanML3D dataset since (1) we found their motion representation contains redundant information, and (2) an extra optimization step would be required to convert their motion representations to the original SMPL format resulting in a degradation in the quality and speed of motion generation in the SMPL format. Similar to HumanML3D, we double the size of the dataset by mirroring the motions and editing the textual description accordingly, resulting in 26850 motion examples in the training set. Additionally, we benefit from the GTA-Human dataset~\cite{gtahuman} built from an open-world action game providing SMPL annotations overlaid on the virtual humans for 20000 motion samples. This dataset does not have any textual annotation, thus we use it for an unconditional training of our model simultaneous to its text-conditional training. 

Since the existing mocap datasets are limited in the number of samples and their diversity, we also collected a large-scale dataset of in-the-wild human poses from image datasets as follows. 
\paragraph{Large-scale Text Pseudo-Pose (TPP) dataset.} Similar to~\cite{azadi2023text}, we have collected a dataset containing 35M pairs of human poses and their text descriptions from a few large-scale image-text datasets. We processed all images by running Detectron2 keypoint detector to find images with a single human, then extracted their 3D pseudo-pose SMPL annotations using a
pre-trained PyMAF-X model~\cite{pymafx}. This large-scale data overcomes the limitations of the existing mocap datasets by
providing a wide variety of human poses and a huge number of
(text, 3D pose) sample pairs. We do not extract any face or hand expressions from the aforementioned datasets.

\subsection{Diffusion Models Background}
Diffusion models are a class of generative model that convert Gaussian noise into samples from a learned distribution by iterative denoising. The forward process is a Gaussian noising process following a Markove chain, $\{x_t\}_{t=0}^T$ where $x_0$ is drawn from the data distribution, in our case a set of 3D SMPL body poses. We can express the forward process as 
$q(x_t | x_{t-1}) = \mathcal{N}(\sqrt{\alpha_t}x_{t-1},(1-\alpha_t)\mathbf{I})$ where $\alpha_t \in (0,1)$. As is standard in training diffusion models, we learn to reverse the forward process by minimizing a noise-prediction loss \cite{ddpms}:
 \begin{equation}
    \mathcal{L}_{\textrm{simple}} = \mathbb{E}_{\epsilon\sim\mathcal{N}(0, \mathbf{I}),t\sim U[1,T]}\left[\left\|\epsilon_t-\epsilon_\theta(x_t, c, t)\right\|_2^2\right]
\label{eqn: loss}
\end{equation}
where $c$ is the optional conditioning for the diffusion model, which in our cases are text descriptions.

In addition to this {\textit{simple}} loss, we follow ~\cite{nichol2021improved,ramesh2022dalle2} and optimize the {\textit{hybrid}} loss, which also uses a loss $L_{vlb}$ that adds a constraint on the estimated variational lower bound (VLB). The $L_{vlb}$ term is applied the same way as in~\cite{nichol2021improved}. We follow \cite{ImagenVid, progdistill} and parameterize our models using $v$-parameterization $(v_t = \alpha_t \epsilon -\sigma_t x)$ rather than predicting $\epsilon$ or $x$.

\subsection{Make-An-Animation}

We represent an avatar body pose, $P$, via 3D SMPL body parameters and an avatar motion as a sequence of body poses for $N$ frames, $[P_1, P_2, \cdots, P_N]$. We use 6D continuous SMPL representation for 21 body joints and the root orient. We additionally represent the global position per frame via a 3D vector indicating the avatar's position in each of the $x, y, z$ dimensions, resulting in each $P_i \in \mathbb{R}^{135} $.  
Make-An-Animation is built from three major components: (1) a pre-trained language model trained on a large-scale language data (T5-XXL~\cite{t5_2020}), (2) a text-to-3D diffusion based pose generation model trained on our collected TPP dataset, and (3) a set of temporal convolution and attention layers extending the static pose generation model to the temporal dimension for motion generation and capturing the dependencies between the frames. 

We follow~\cite{ImagenVid} and use v-prediction parameterization $(v_t = \alpha_t \epsilon -\sigma_t x)$ to train our pose and motion diffusion models for numerical stability. Our denoising model in the above U-Net architecture operates on all motion frames at the same time resulting in a higher temporal coherence in the generated motions compared to the autoregressive frameworks~\cite{tevet2022mdm} and does not require any specific loss on the motion velocity for a smooth motion synthesis.
We also benefit from classifier free guidance~\cite{cfg} at inference by conditioning our model on a null text during training for $10\%$ of the time.

\begin{figure*}
  \centering \includegraphics[width=\textwidth]{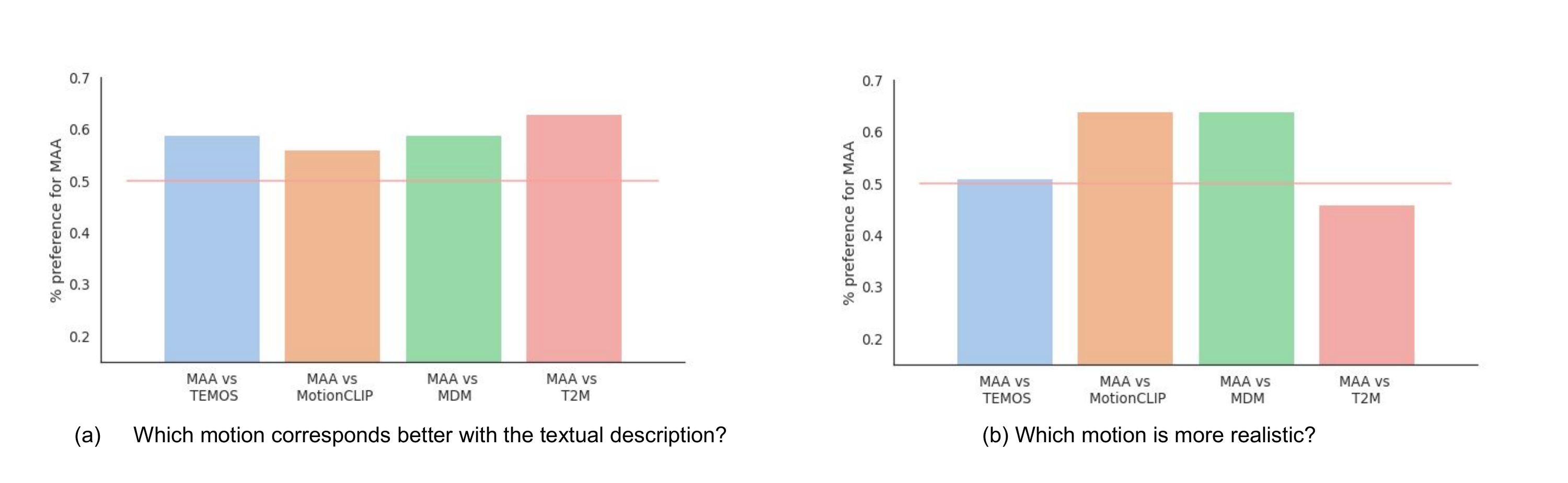}
  \caption{\textbf{Human Evaluation.} Here we compare each baseline against our model, MAA, in terms of text alignment and motion realism.
The results are reported as the percentage of the majority vote raters that
preferred our method to each baseline on our curated 400 prompts set.} 
  \label{fig:amt}
\end{figure*}

\input{Tables/auto.tex}

\begin{figure}
  \centering \includegraphics[width=0.49\textwidth]{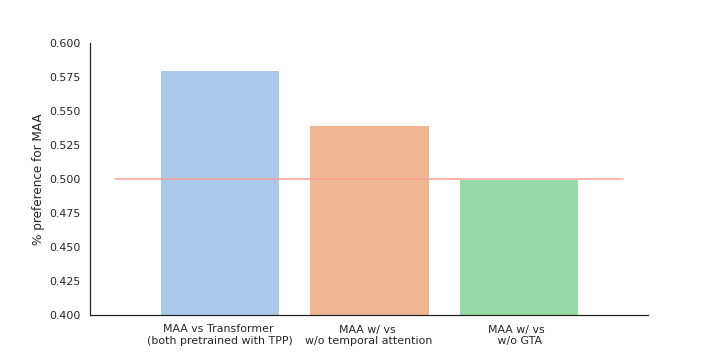}
  \caption{\textbf{Ablation Study.} Here we compare each ablated model against our model, MAA, in terms text alignment and motion realism.
The results are reported as the percentage of the majority vote raters that
preferred our method to each baseline on our curated 400 prompts set.}
  \label{fig:ablation}
\end{figure}

\subsubsection{Text-to-3D Pose Generation}
We train a U-Net based diffusion model to generate 3D SMPL pose parameters conditioning on text embeddings from a large frozen language model~\cite{t5_2020}. Inspired by the image and video generation models~\cite{MAV}, we reshape the pose inputs to $B\times C\times 1\times 1$ with $B$ and $C$ indicating batch size and channel dimension of 135, respectively. Our U-Net model is built from a sequence of (1) Residual blocks with 1x1 2D-convolution layers conditioned on the diffusion time-step embedding and text embedding, and (2) Attention Blocks attending to the textual information and the diffusion time-step.
We train this network on our large-scale TPP dataset where we set translation parameters to zero in all training examples, i.e., human at the center of the scene.

\subsubsection{Temporal and Attention Layers}
In order to expand the aforementioned pose generation model to learn the temporal dimension for motion generation, we modify the convolutional and attention layers of the U-Net as follows.

We reshape the input motion sequence to a tensor of shape $B \times C \times N \times 1 \times 1$, where $C$ is the length of the pose representation and $N$ is the number of frames. 
Inspired by~\cite{MAV, ImagenVid}, we stack a 1D temporal convolution layer following each 1x1 2D-convolution. This allows us to train the new 1D temporal layers from scratch while loading the pre-trained convolution weights from the pose generation model. We set the kernel size of these temporal convolution layers to 3 opposed to the unit kernel size of the 2D convolutions. 
We apply a similar dimension decomposition strategy to the attention layers, where we stack a newly initialized temporal attention layer to each pre-trained attention block from the pose generation network. For the temporal attention layers we utilize a Rotary Position Embedding~\cite{roformer}.

% \subsubsection{Training}

\section{Experiments}
We compare the performance of our motion generation model with
the existing state-of-the-art text-to-human-motion generation models through human evaluation on 400 crowd-sourced prompts and automatic metrics on the HumanML3D test set.

\subsection{Automatic Metrics}
We perform an automatic evaluation on the HumanML3D test set in terms of Fr\'echet Distance (FID), R-Precision and Diversity scores. FID measures both diversity and quality of the samples comparing its distribution with the ground truth test set while R-Precision measures faithfulness between each generated motion and its input prompt. Diversity measures the variance of the generated motions across all descriptions. To compute the above scores, a motion encoder was jointly trained with a text encoder via a contrastive loss by~\cite{guo2022t2m}. We use the same encoder to extract text and motion embeddings. Since this model has been trained and optimized on the 263-dimensional pose representation vector, we transform both the ground-truth SMPL data and our generated samples similarly following the transformations provided by~\cite{guo2020a2m} before computing the above metrics. We upsample our generations to 196 frames and keep that fixed for the ground truth samples and our outputs while computing the above metrics. Opposed to our baselines, we have not optimized our model on the 263-dimensional motion representation, but achieve a comparable performance in all metrics as reported in Table~\ref{tab:auto}.

\subsection{Human Evaluation}
\begin{figure*}
  \centering \includegraphics[width=\textwidth]{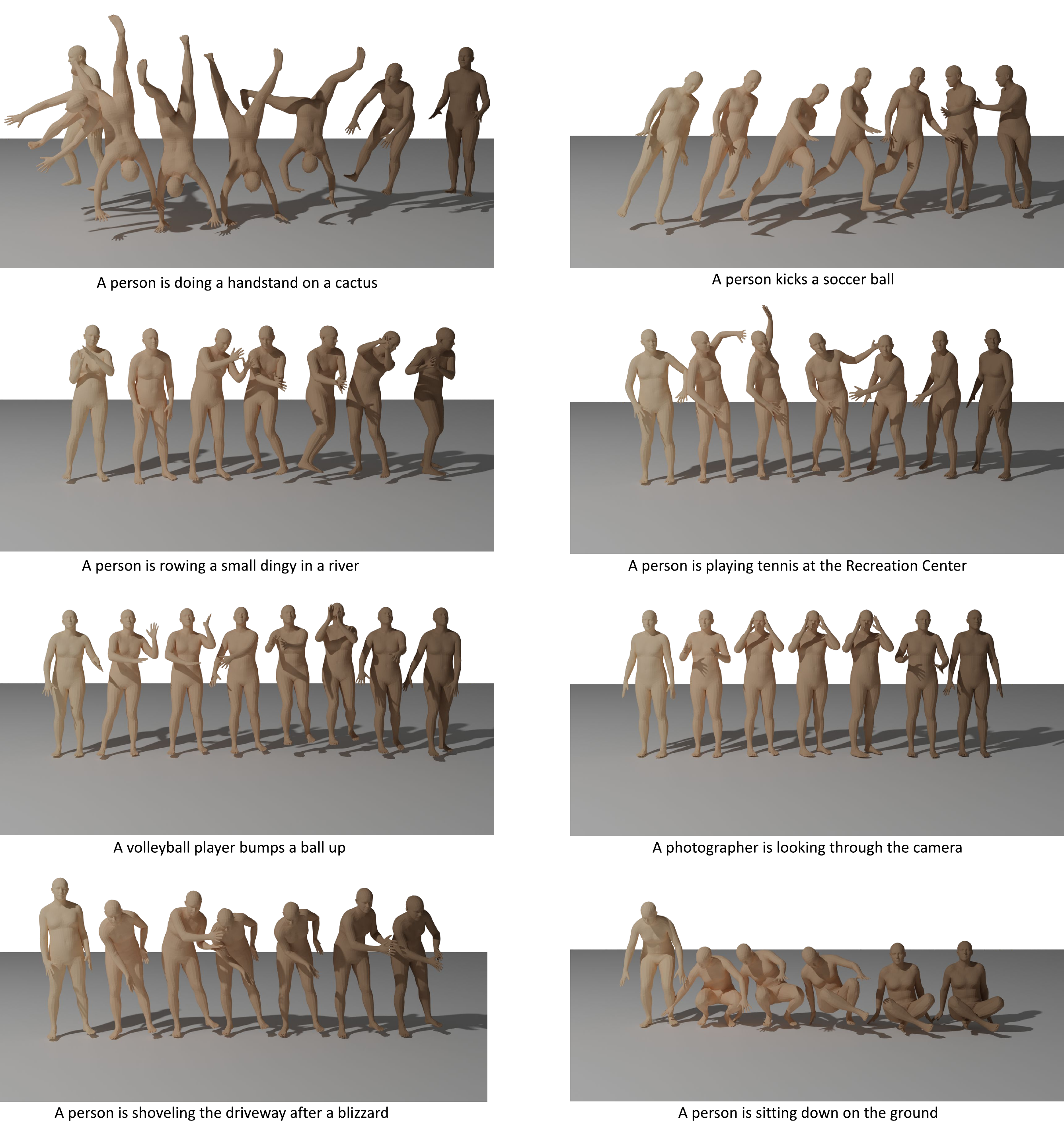}
  \caption{Samples generated by Make-An-Animation for text conditional motion generation. The lighting of the body models represents progress across time. Darker color indicates later frames in the sequence. For a better visualization, frames are distributed horizontally. }
  \label{fig:samples}
\end{figure*}

To evaluate the generalization ability of our model in synthesizing more challenging human poses and motions, we collected an evaluation set from Amazon Mechanical Turk (AMT) that consists of 400 prompts. We asked annotators for prompts that described an action along with some context of the scene. We filtered out prompts based on ethical concerns to remove any references to children or NSFW content. These prompts were selected without generating any poses or images for them, and were kept fixed for all of our evaluations. We generated animated poses from all models with no translation to measure the correctness and quality of the body pose and motion. 
We show raters a text description and the corresponding rendered
avatars from two models and ask them which output better matches
the input text. For each comparison, we use the majority vote from
7 different annotators on 400 prompts as the final result, reported in Figure~\ref{fig:amt}. This study confirms the superiority of our model and the impact of our large-scale TPP dataset in human motion generation and overcoming the limitations of the mocap datasets. A few motion samples generated by our model are shown in Figure~\ref{fig:samples}.

\begin{figure}
  \centering \includegraphics[width=0.4\textwidth]{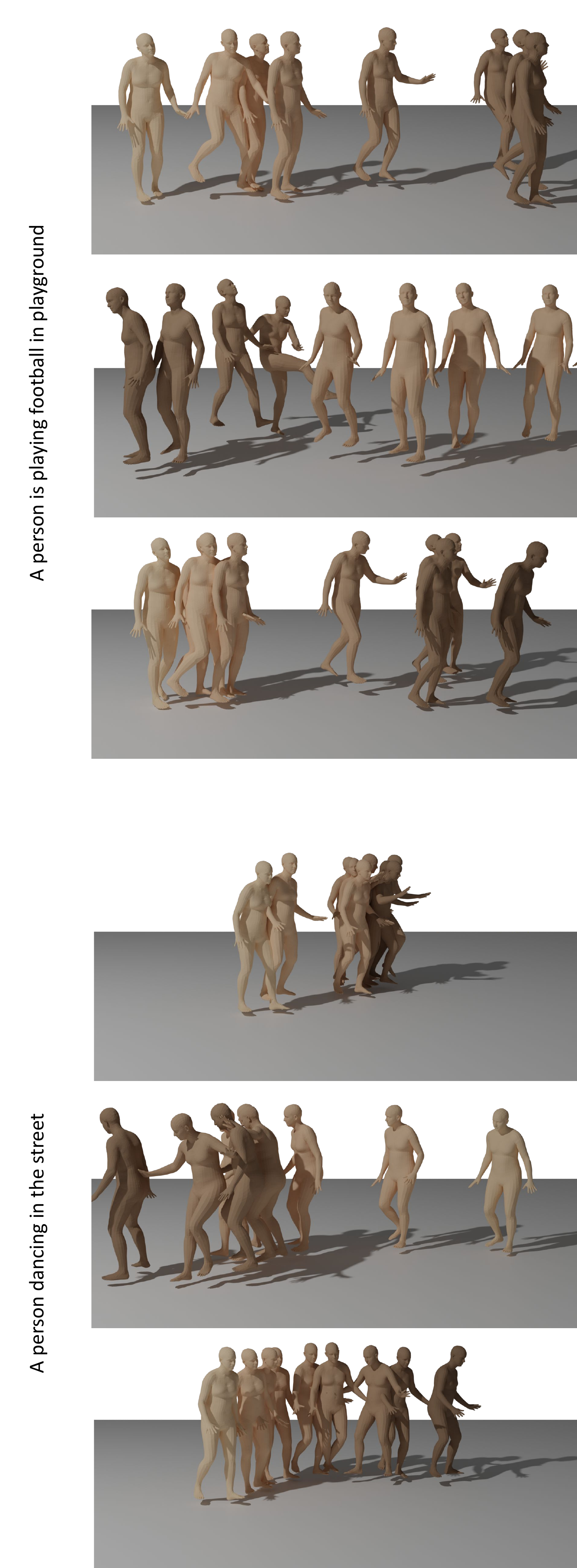}
  \caption{Diverse samples generated by Make-An-Animation for text conditional motion generation. The lighting of the body models represents progress across time. Darker color indicates later frames in the sequence. For a better visualization, frames are down-sampled along the temporal dimension and distributed horizontally. }
  \label{fig:diversity}
\end{figure}

\begin{figure*}
  \centering \includegraphics[width=\textwidth]{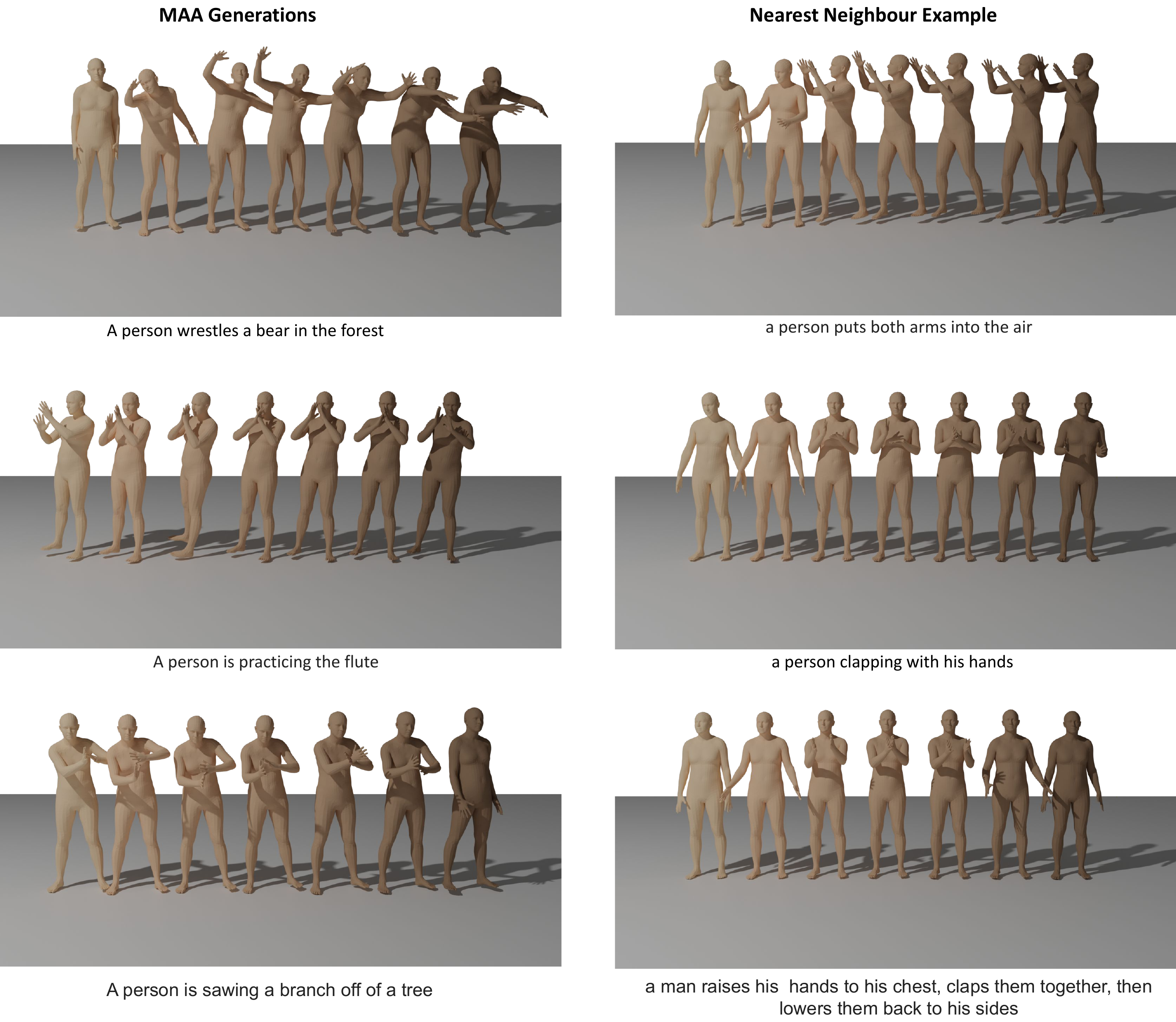}
  \caption{Left: samples generated by Make-An-Animation for text conditional motion generation. Right: The nearest neighbor example for each of the generated samples found from the mocap training set based on motion and text clip similarity scores. The lighting of the body models represents progress across time. Darker color indicates later frames in the sequence. For a better visualization, frames are distributed horizontally. } 
  \label{fig:fig_nn}
\end{figure*}
\subsection{Ablation Studies}
To disentangle and study the impact of our U-Net architecture versus the impact on our TPP dataset on our results, we implemented and trained an alternative diffusion model as a decoder-only transformer with a causal attention mask to generate 3D pose reprsentations from text. This model operates on a sequence of tokenized captions and their text embeddings, the diffusion time-step embedding, the noised body pose and root orient representations, and two final pose and orientation queries to predict
the unnoised pose and root orientation, respectively. We pre-trained this model on our TPP dataset, then expanded the model for motion synthesis via adding three additional tokens per frame for pose, root orient, and translation paramters. We trained this model on the mocap datasets described in Sec~\ref{sec:mocap} and compared its performance with our U-Net model via Amazon Mechanical Turk. As a result, $58\%$ of the raters preferred the generations from our U-Net architecture. This experiment confirms the superiority of the U-Net model versus its auto-regressive alternatives. We also ablated the impact of temporal attention modules on the final quality of the samples resulting in $54\%$ of the users preferring samples from a U-Net with temporal attention. 

Training with the GTA data unconditionally made the convergence of the model faster although it didn't impact the samples' quality after convergence. 

Additionally, we experimented with training our U-Net model on the mocap training set from scratch and without pre-training. However, we observed stability and convergence issues in this setting even with various learning rates and warming up updates. The pre-training stage on the large-scale dataset was therefore a stable solution resulting in high quality motion samples in the end. All ablation results are summarized in Figure~\ref{fig:ablation}.

\subsection{Qualitative Studies}
In Figure~\ref{fig:fig_nn}, we visualize the generalization capability of our model in synthesizing novel motions. We investigate the existence of similar prompts or motions in the training data for each synthesized motion via a nearest neighbor search. We extract the motion embedding and text embedding for each sample in our test set of prompts as well as our mocap training data using a pre-trained motionCLIP model~\cite{tevet2022motionclip, kim2022flame} that has learned a joint motion and language embedding space. For each synthesized motion, we find the top 6 motions and prompts from the training data with the highest cosine similarity scores. As shown in the figure, our model has generalized to new motions that did not exist in the training data. In addition, Figure~\ref{fig:diversity} illustrates multiple diverse motions generated by our model for different text prompt.

\section{Conclusion}
Current text-to-motion models are hindered by relatively small-scale motion capture datasets, as evident by their poor performance on more diverse, in-the-wild prompts. Motivated by this shortcoming, we present Make-An-Animation, a human motion generation model trained on a collection of large-scale static pseudo-pose and motion capture data. As we demonstrate through our ablations and human studies comparing to prior works, pre-training on the large-scale TPP dataset significantly improves performance on captions outside of the distribution of motion capture data. Furthermore, our novel U-Net architecture enables a seamless transition between static pose pre-training and dynamic pose (i.e., motion) fine-tuning. Altogether, this work paves the way to leverage large-scale image and video datasets for learning generation of human 3D pose parameters.

\newpage

{\small
\bibliographystyle{ieee_fullname}
\bibliography{egbib}
}

\end{document}

%% file: Tables/auto.tex
\begin{table}[t]
    \begin{center}
        \setlength{\tabcolsep}{2pt}
            \begin{tabular}{lccc}
                \toprule
                \textbf{Method} & 
                R-Precision $\uparrow$ &
                FID $\downarrow$ &
                Diversity $\uparrow$ \\
                \midrule
                 Real &
                $0.797^{\pm 0.002}$ &
                 $0.002^{\pm 0.002}$&
                 $9.5^{\pm 0.65}$ \\
                 \midrule
                J2LP[\cite{ahuja2019jl2p}] &
                 $0.486^{\pm .002}$&
                 $11.02^{\pm .046}$&
                 $7.676^{\pm .058}$ \\

                 MDM[\cite{tevet2022mdm}] &
                 $0.611^{\pm.007}$&
                 $0.544^{\pm .044}$&
                 $9.559^{\pm .086}$ \\
                  T2M[\cite{guo2022t2m}] &
                 $0.740^{\pm .003} $&
                 $1.067^{\pm .002}$&
                 $9.188^{\pm .002}$ \\
                 \midrule
                MAA &
                 $0.6755 ^{\pm .002} $&
                 $0.7740 ^{\pm  0.007 }$&
                 $8.23^{\pm 0.064}$ \\

                \bottomrule
            \end{tabular}
    \caption{\textbf{ Quantitative results on the HumanML3D test set.} We run our evaluation 20 times and $\pm$ indicates the $95\%$
confidence interval.}
    \vspace{-2em}
    \label{tab:auto}
    \end{center}
\end{table}